%% file: root.tex
\documentclass[letterpaper, 10 pt, conference]{ieeeconf}  % Comment this line out if you need a4paper

\IEEEoverridecommandlockouts                              % This command is only needed if 
\usepackage{gensymb}
\usepackage{xcolor}
\usepackage{amsmath}
\usepackage{graphicx}
\usepackage{cite}
\usepackage[caption=false]{subfig}
\usepackage{prettyref}
\newrefformat{fig}{Fig.~\ref{#1}}
\newrefformat{sec}{Sec.~\ref{#1}}
\newrefformat{tab}{Tab.~\ref{#1}}
\newrefformat{algo}{Algo.~\ref{#1}}
\newrefformat{eq}{(\ref{#1})}
\title{\LARGE \bf
Polymer-Based Self-Calibrated Optical Fiber Tactile Sensor
}

\author{Wentao Chen$^{1}$, Youcan Yan$^{2}$, Zeqing Zhang$^{1}$, Lei Yang$^{3,1}$ and Jia Pan$^{1}$% <-this % stops a space
\thanks{$^{1}$Department of Computer Science,
        The University of Hong Kong, Pokfulam, Hong Kong}%
\thanks{$^{2}$CNRS-University of Montpellier, LIRMM, Interactive Digital Humans group, Montpellier, France}%
\thanks{$^{3}$Centre for Transformative Garment Production, Hong Kong}%
}

\begin{document}

\maketitle
\thispagestyle{empty}
\pagestyle{empty}

%%%%%%%%%%%%%%%%%%%%%%%%%%%%%%%%%%%%%%%%%%%%%%%%%%%%%%%%%%%%%%%%%%%%%%%%%%%%%%%%
\begin{abstract}

Human skin can accurately sense the self-decoupled normal and shear forces when in contact with objects of different sizes. Although there exist many soft and conformable tactile sensors on robotic applications able to decouple the normal force and shear forces, the impact of the size of object in contact on the force calibration model has been commonly ignored. Here, using the principle that contact force can be derived from the light power loss in the soft optical fiber core, we present a soft tactile sensor that decouples normal and shear forces and calibrates the measurement results based on the object size, by designing a two-layered weaved polymer-based optical fiber anisotropic structure embedded in a soft elastomer. Based on the anisotropic response of optical fibers, we developed a linear calibration algorithm to simultaneously measure the size of the contact object and the decoupled normal and shear forces calibrated the object size. By calibrating the sensor at the robotic arm tip, we show that robots can reconstruct the force vector at an average accuracy of \textbf{0.15N} for normal forces, \textbf{0.17N} for shear forces in X-axis , and \textbf{0.18N} for shear forces in Y-axis, within the sensing range of \textbf{0-2N} in all directions, and the average accuracy of object size measurement of 0.4mm, within the test indenter diameter range of \textbf{5-12mm}.

\end{abstract}

\input{sections/introduction}

\input{sections/methodology}

\input{sections/experiments}

\input{sections/conclusion}

\addtolength{\textheight}{-12cm}   % This command serves to balance the column lengths

%%%%%%%%%%%%%%%%%%%%%%%%%%%%%%%%%%%%%%%%%%%%%%%%%%%%%%%%%%%%%%%%%%%%%%%%%%%%%%%%

%%%%%%%%%%%%%%%%%%%%%%%%%%%%%%%%%%%%%%%%%%%%%%%%%%%%%%%%%%%%%%%%%%%%%%%%%%%%%%%%

%%%%%%%%%%%%%%%%%%%%%%%%%%%%%%%%%%%%%%%%%%%%%%%%%%%%%%%%%%%%%%%%%%%%%%%%%%%%%%%%

%\begin{thebibliography}{99}
\bibliographystyle{IEEEtran}
\bibliography{ref}

%\end{thebibliography}

\end{document}

%% file: sections/introduction.tex
\section{Introduction}\label{sec:intro}
\begin{figure*}[t]
    \centering
    \includegraphics[width=2\columnwidth]{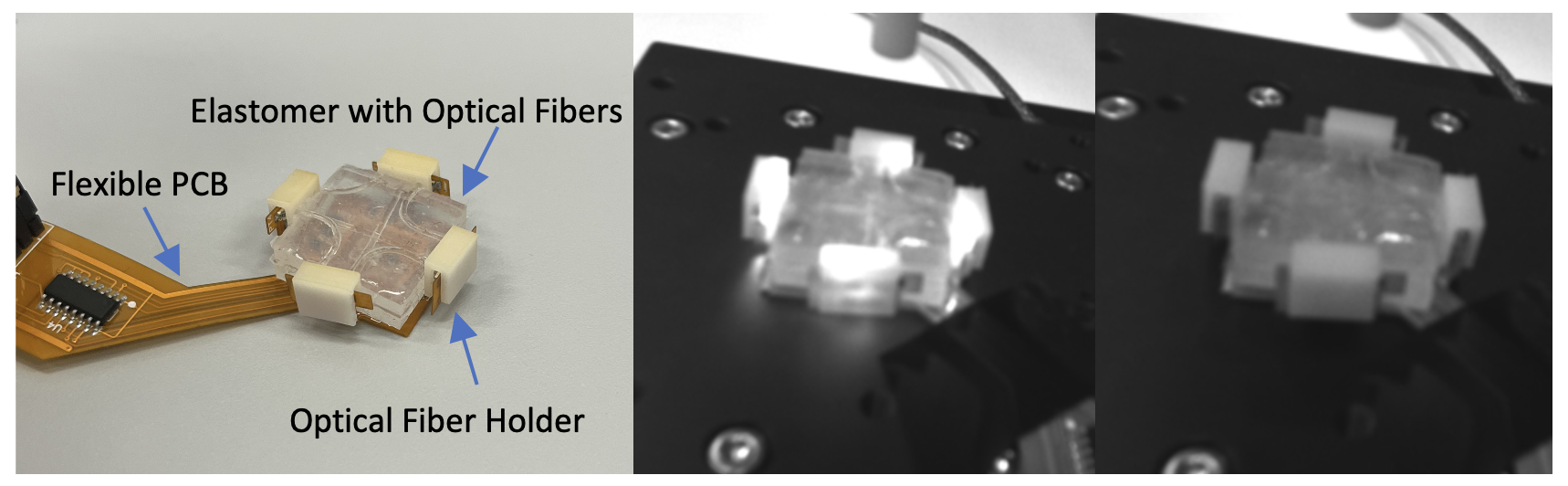}
    \caption{The overall configuration of the sensor. The sensor consists of an elastomer cladding with two embedded layers of optical fiber cores. A flexible PCB with integrated Infrared radiation (IR) light sources (LEDs) and receivers (photodiodes, PD) is wrapped around the elastomer, connected by four optical fiber holders (left picture). The middle and right pictures are taken with an Intel RealSense IR camera, which corresponds to the IR-on and -off status within the optical fibers.}
    \label{fig:fig1}
\end{figure*}
%Human skin is amazingly good at precepting the sense of touch when human interacts with the environment. Inspired by this, over the years, artificial tactile sensors have been wide in a variety of robotic dexterous manipulation tasks requiring fine tactile feedback. Although progresses on the artificial tactile sensor has been made, achieving fine tactile feedback and dexterous manipulation continues to be a challenge. One of the reasons is that the robots lack flexible artificial skin that achieves the same perception function as humans. Thanks to the dense distribution of different types of fast and slow-adaptive mechanoreceptors, human skin can perceive the continuous deformation field induced by the external contact wrench and simultaneously extract the information of the exerted force from the deformation of the deformation field. Hence, designing a flexible tactile sensor with a biomimetic approach that can achieve the simultaneous perception of the deformation field and tactile force is important for improving robot capabilities and would expand its application in the physical human-robot interaction (pHRI) environment.

Human skin can perceive the deformation field induced by the external contact wrench and simultaneously extract the information of the exerted force and object size from the deformation field\cite{johansson1983tactile}. This is one of the key enablers that allow human hands to perform a wide range of manipulation tasks. To match the human-level dexterity in manipulation, it is, therefore, crucial to equip modern robotic grippers with artificial tactile sensors that can perceive rich contact information for achieving the manipulation tasks.

Recently, many artificial tactile sensors that have the ability to measure the deformation field (with information on force and object size) on the surface have been proposed. One major category focuses on using a vision-based sensor to capture the normal and tangential movement of markers embedded underneath the elastomer surface \cite{yuan2017gelsight,guo2016measurement,sui2021incipient,du2021high,zhang2022multidimensional,ma2019dense,lin2020curvature,wang2021gelsight}. Using the photometric stereo algorithm, the depth map of surface deformation on the sensing surface can be reconstructed or inferred\cite{yuan2017gelsight}.  Within the consecutive frames of images of the sensing surface, the motion of the markers enables the calculation of  normal and tangential displacements or each subsection of the surface, which can then be used to infer contact force using inverse finite element methods (iFEM)\cite{sui2021incipient,du2021high,ma2019dense} or machine learning methods \cite{guo2016measurement,zhang2022multidimensional}. Besides the camera-based sensors, there are other non-camera-based sensors that also measures deformation field, such as BioTac that uses fluidic channel electrodes \cite{narang2021sim}, electrical impedance tomography (EIT)-based tactile sensor \cite{park2022biomimetic} and magnetic film-based sensor \cite{yan2021soft}. These sensors also use deep learning (variational autoencoder in \cite{narang2021sim}, the deep neural network in \cite{park2022biomimetic}), and iFEM to reconstruct the deformation field.

However, there are several limitations to the sensors mentioned above. Firstly, the relationship between the marker motions and the contact force is not always linear and can be easily affected by the contact object geometry, which will lead to the demand for a large dataset that contains the indentation profile of different shapes\cite{guo2016measurement}. Also, the accuracy of this mechanism is highly sensitive to the parameters of the camera and requires image adaptation under different conditions\cite{du2021high,zhang2022multidimensional}. On the other hand, the measurements of the displacement field and force field are highly coupled, requiring either image processing or assumption of material property in iFEM for force field estimation from displacement field\cite{ma2019dense,wang2021gelsight}.   Lastly, the aforementioned studies all rely on bulky and sophisticated hardware, which limits their applications where conformality and structural compactness are required. 
%\yl{[Comments: It is better if we can enumerate the limitations, such as first..., second ..., third ..... It is better if we can first summarize the limitations, each with a few words (giving it a name), and then we explain why this limitation exists or extend our descriptions. Then, we provide some intuitions or design rationales to give the readers a rough grasp of why our proposed solution works better than the previous methods -- how their limitations are addressed by our proposed method. We can present our advantages in an order that the limitations of the previous works are presented. So the readers can quickly match our advantages to their limitations.]}

To tackle the existing challenges in tactile deformation field measurement, we present a tactile sensor based on woven polymer-based optical fibers. Inspired by \cite{cao2022polymer} and  \cite{zhou2022conformable} that arrange multiple optical fibers in different orientations to decouple the normal and shear forces, we arranged the optical fibers in a two-layer way to decouple the normal force, shear force, and the indenter size simultaneously. Different from the previous sensors, our tactile sensor uses optical fibers of different shapes to create an anisotropic response in different directions of the elastomer. This directly gives a mapping from the optical fiber signals to the tactile force and contact object size without relying on a large amount of data or post-processing, as resolved by the combination of intertwining optical fibers within the sensing volume. This also relieves the force measurement result from being affected by the size of the contact object with the self-calibration algorithm, which does not exist in the previous soft optical fiber-based tactile sensor. Based on the linear relationship between the light intensity loss within the fiber and the deformation length of the fiber \cite{bai2020stretchable,zhao2016optoelectronically}, we propose a linear self-calibration algorithm to solve for the contact object size and then calibrate the decoupled normal and shear forces based on the object size from the readings of the fiber network. The measurement result was compared with the ground truth data and achieved good accuracy. 

To summarize, our key contributions are the following:
\begin{enumerate}
    \item A tactile sensor design based on two layers of weaved optical fibers produces an anisotropic response to deformation in different directions. 
    \item A linear self-calibration algorithm that directly maps from optical fiber signals to the contact object size and the tactile force calibrated by the contact object size.
    \item Demonstration of accurate measurement result based on the proposed sensor hardware design and the calibration algorithm. 
\end{enumerate}
 
%\yl{I think we explicitly explain the major idea why our weaved optical fiber sensor can measure the continuous field. And how accurate it is. And may be how efficient it is.}

%\textcolor{red}{1. the core limitations of the previous researches 2. each contribution corresponds to each limitation 3. talk about the new layer structures, why use that}

% \begin{figure*}[t]
%     \centering
%     \includegraphics[width=2\columnwidth]{Picture1.png}
%     \caption{The overall configuration of the sensor. The sensor consist of a elastomer cladding with two embedded layers of optical fiber cores. A flexible PCB with integrated Infrared radiation (IR) light sources and receivers is wrapped around the elastomer, connected by four optical fiber holders.}
%     \label{fig:fig1}
% \end{figure*}

%% file: sections/methodology.tex
\section{Sensing Principle and Hardware} \label{sec:principle}

\subsection{Sensor Design and Working Principles}
To realize the anisotropic response of the optical fiber to deformations in different directions, we propose the following sensor design. The proposed tactile sensor is composed of two layers of soft optical fiber and a printed circuit board (PCB) for the emission and detection of light. A soft elastomer encloses the soft optical fibers in the layout as shown in Fig.~\ref{fig:fig1}, working as the cladding. Based on the principle of total internal reflection \cite{li1996measurement}, where the refractive index of cores $n_{core} \approx 1.48$ is higher than that of the cladding $n_{clad} \approx 1.41$, the optical ray will be entirely constrained within the fiber cores and the loss is negligible. Due to the property that the force on the surface of soft material will also induce strains deep within the material, different combinations of the normal forces and shear force will incur different strain distributions within the elastomer. When forces are exerted on the elastomer surface, the deformation induced within the elastomer will also cause the change of shape of the core fibers and leads to the power loss of light within the core. On the other hand, the indentation depth (correlated to the normal force) and the geometric size of the object will also affect the strain within the elastomer.

\begin{figure}[t]
    \centering
    \includegraphics[width=\columnwidth]{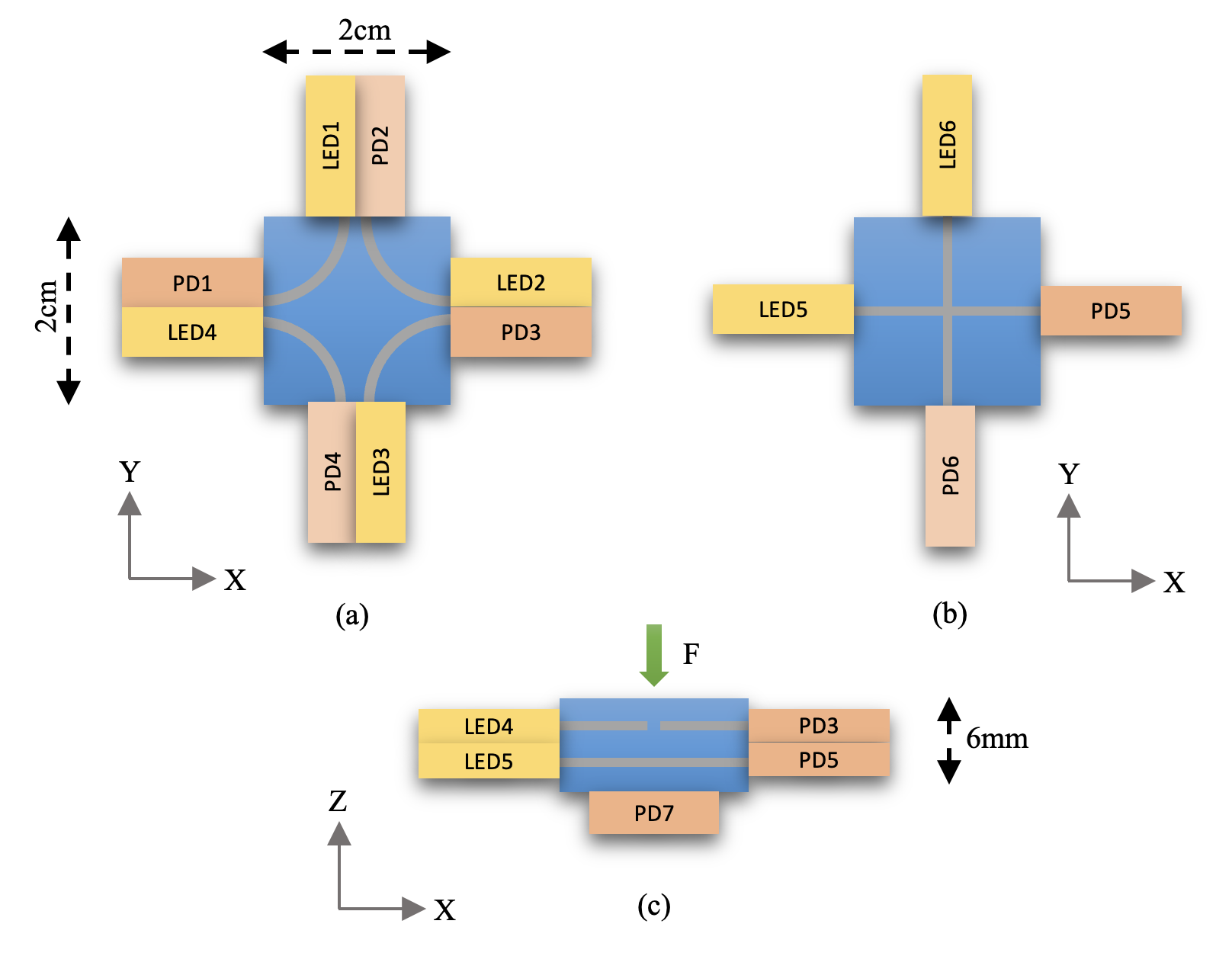}
    \caption{The design of the proposed tactile sensor. (a) and (b) illustrate the upper and lower layers of the optical fiber structure, with the grey line denoting the optical fibers. (c) shows the lateral sectioned view of the structure within the elastomer. The peripheral IR light-emitting LEDs and IR-sensing PDs are as shown. Note: the dimensions in the picture are not strictly scaled.}
    \label{fig2}
\end{figure}

There have been previous works as listed in Sec.~\ref{sec:intro} that demonstrated the feasibility of soft waveguide-based force sensors. Here, we are refining the force sensing capability of the soft waveguide-based tactile sensor by taking the size of contact object into consideration. Based on the hypothesis that the force magnitude and the object size will affect the length of deformed section of the fiber core within the elastomer, the power loss of light within the fiber under external load, i.e. the sensor response, contains the coupled force and object size information and should be decoupled. Therefore, we propose the two-layered structure of soft polyurethane (PU) optical fiber for validating this hypothesis in the next section, as shown in Fig.~\ref{fig2}. For each fiber, one end is connected to an IR-emitting LED chip, and the other end to an IR photodiode (PD) chip. 

The differences in the shape of the optical fibers are intended to create the anisotropic fiber response for extracting decouple force and object size information. The top layer of this structure is composed of four U-shaped optical fiber that is mainly sensitive to the shear load component. The bent portion of the U-fiber is more sensitive to the shear deformation, and the portion of elastomer between the U-fiber and the bottom PCB ensures enough normal support force so that it is less susceptible to normal direction deformation. In contrast, the bottom layer is composed of two vertically crossing straight fibers that are designed to be primarily sensitive to the normal load component, as the optical fiber is assumed to have negligible deformation in the axial direction. The portion of elastomer between the lateral side of the sensor and the straight fiber also ensures that the straight fiber is less susceptible to shear deformation. Besides the PD chips that are attached to the ends of the optical fibers, there is one additional PD chip attached to the bottom of the elastomer, which reads the light power leakage from the bottom layer of optical fibers and is only sensitive to the indentation depth. Under successful validation, the combination of the PD readings on the PCB can be used to obtain the object size and the calibrated force vector simultaneously. 

\subsection{Manufacturing Process}

\begin{figure}[t]
    \centering
    \includegraphics[width=\columnwidth]{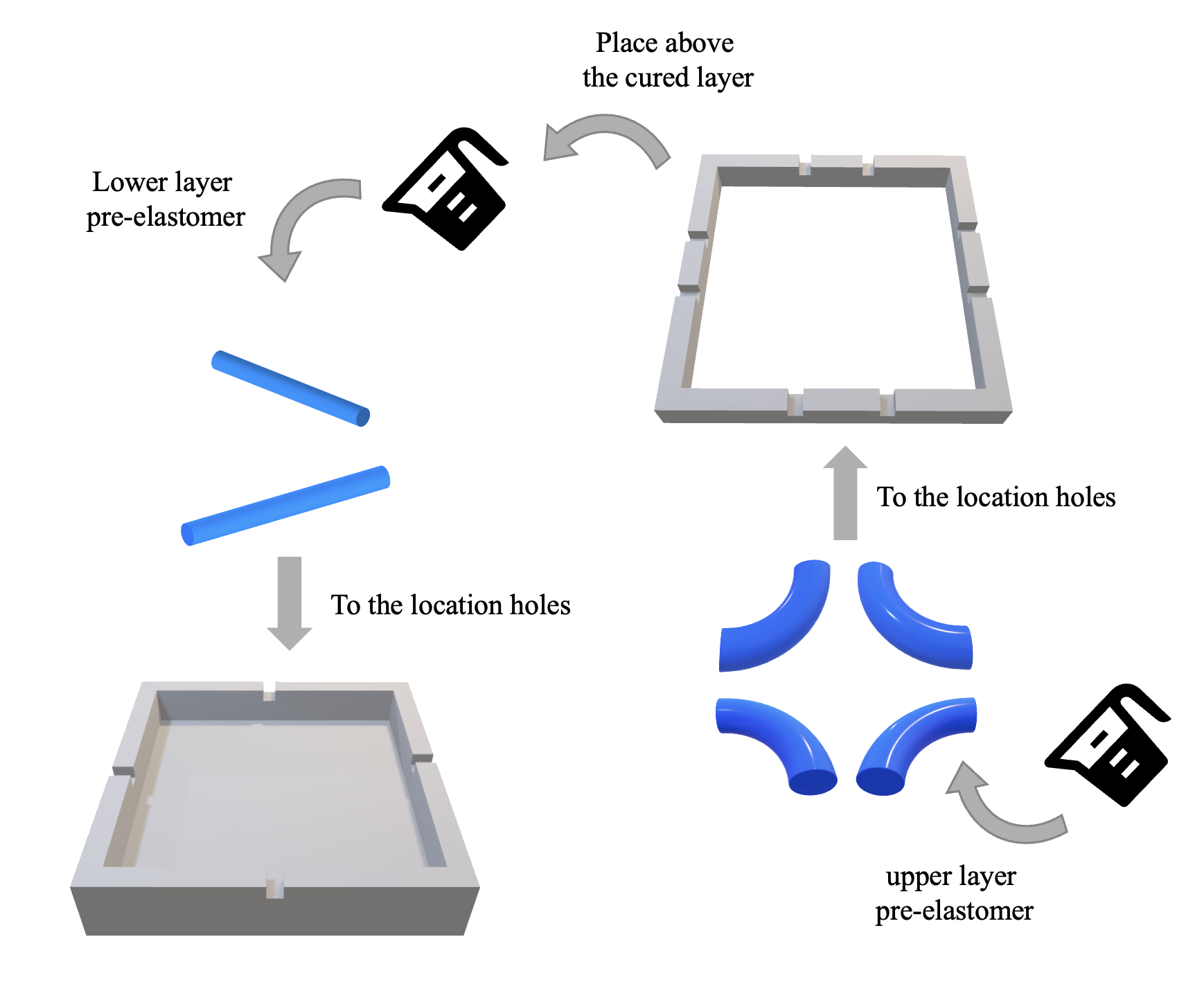}
    \caption{Illustration of the fabrication of the sensor elastomer with the two-layer optical fiber structure. The fabrication steps are proceeded from left to right. When manufacturing each layer, the ends of optical fibers (shown as blue tubes in the picture) are held at the location holes to either be pre-strained or pre-bent, and the pre-elastomer is poured into the mold and cured. The upper layer of the structure is directly cured above the lower layer.}
    \label{fig:3}
\end{figure}

The manufacturing process of the tactile sensor consists of three key steps: topographic molding of the elastomer cladding and optical fiber core, PCB layer manufacturing, and assembly of the elastomer and the PCB layer, as shown in Fig.~\ref{fig:3}. These steps were conducted in the sequence as listed. 
\begin{itemize}
    \item Topographic molding: The upper layer mold and the lower layer mold (Fig.~\ref{fig:3}a) of the elastomer cladding were 3D-printed using the WeNext 8100 SLA polymer resin, by the WeNext 3D Printing service (WeNext Technology Co., Ltd). The location holes for assembling the inner optical fiber cores are located at the middle of each side of the mold, in order to achieve a symmetric arrangement of the optical fibers throughout the interior of the elastomer. The dimension of the position holes is the same size as the diameter of the optical fiber cores (1mm). The dimensions of the elastomer (shown in Fig.~\ref{fig2}) were deliberately chosen in order to balance the transmittance of light within the fiber core and the sensitivity of the sensor. To mold the elastomer cladding with an embedded optical fiber core, we used silicone elastomer Ecoflex 00-50 (Smooth-On Inc.). Parts A and B of Ecoflex 00-50 were mixed at a ratio of 1:1, then the mixture was placed in a vacuum centrifugal mixer for 1 minute, at the speed of 1000 rpm. The mixed pre-elastomer was poured into the lower mold and then placed on a heating platform set at 70\degree C for 15 minutes. After the lower layer is cured, the mold is removed from the cured lower layer, and the upper layer mold is directly placed above the cured lower layer with the edges aligned. The same process is repeated for curing the upper layer of the elastomer. Notice that before the pre-elastomer was poured into the molds, the PU fibers (diameter of 1mm, SHIN-IL Co.) had been pre-strained straight and held at the location holes of the lower layer mold, and have been bent to the shape of a quarter circle and held at the location holes of the upper layer mold. 
    \item PCB layer manufacturing: LEDs (0805 packaged, IR17-21C, Everlight Electronics Co., Ltd.) and PDs (0805 packaged, PT17-21C, Everlight Electronics Co., Ltd.) are the optical resources and signal receptors that are integrated onto a single flexible PCB. The circuit was powered by a 3.3V voltage supply from an Arduino Mega 2560 microcontroller board. The resistance $R_{LED}$ in series with the LED, and the resistance $R_{PD}$ in series with the PD, are of the same values as in \cite{zhou2022conformable}, in order to adjust the luminescence intensity of the LED and the voltage readout of the PD to a level that makes the sensor sensitive enough to the external loading. 
    \item Assembly of the elastomer and the PCB: The combination of the elastomer and the  flexible PCB leads to complete assembly of the tactile sensor. The extruding parts of the flexible PCB, containing the LED-PD pairs at the sides of the elastomer, were bent at 90\degree, facing the lateral sides of the elastomer. To tightly attach the flexible PCB to the elastomer while guaranteeing that the ends of the fiber cores are properly aligned to the LEDs and PDs, four optical fiber holders (as shown in \prettyref{fig:fig1}) were 3D-printed with location holes for the fiber ends and the LEDs and PDs on the flexible PCB. The optical fiber holders connect the elastomer to the four extruding parts of the flexible tightly, so that the sensor can stably output signals under large deformation. 
    
\end{itemize}

\begin{figure}[th]
    \centering
    \includegraphics[width=1.05\columnwidth]{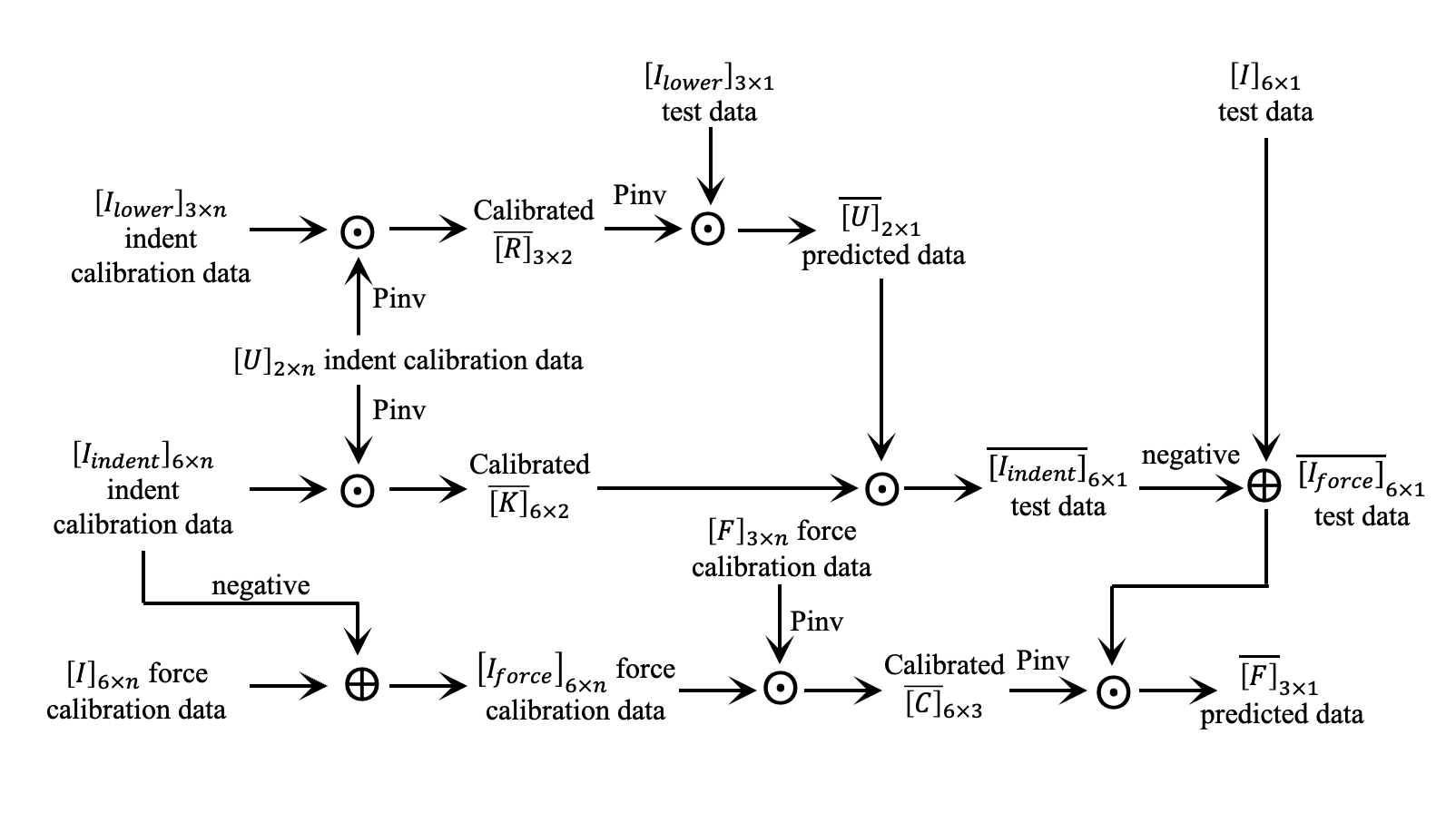}
    \caption{The illustrative diagram of the proposed self-calibration algorithm. The $\bigodot$ represents the dot product of matrices, $\bigoplus$ represents the summation of matrices, 'Pinv' presents the Moore-Penrose pseudoinverse of the matrix, and 'negative' represents the negative of the matrix.}
    \label{fig:8}
\end{figure}

\section{Data Acquisition}

\begin{figure*}[t]
    \centering
    \subfloat[Diagram of the data acquisition and calibration process.]{\includegraphics[width = 1.1\columnwidth]{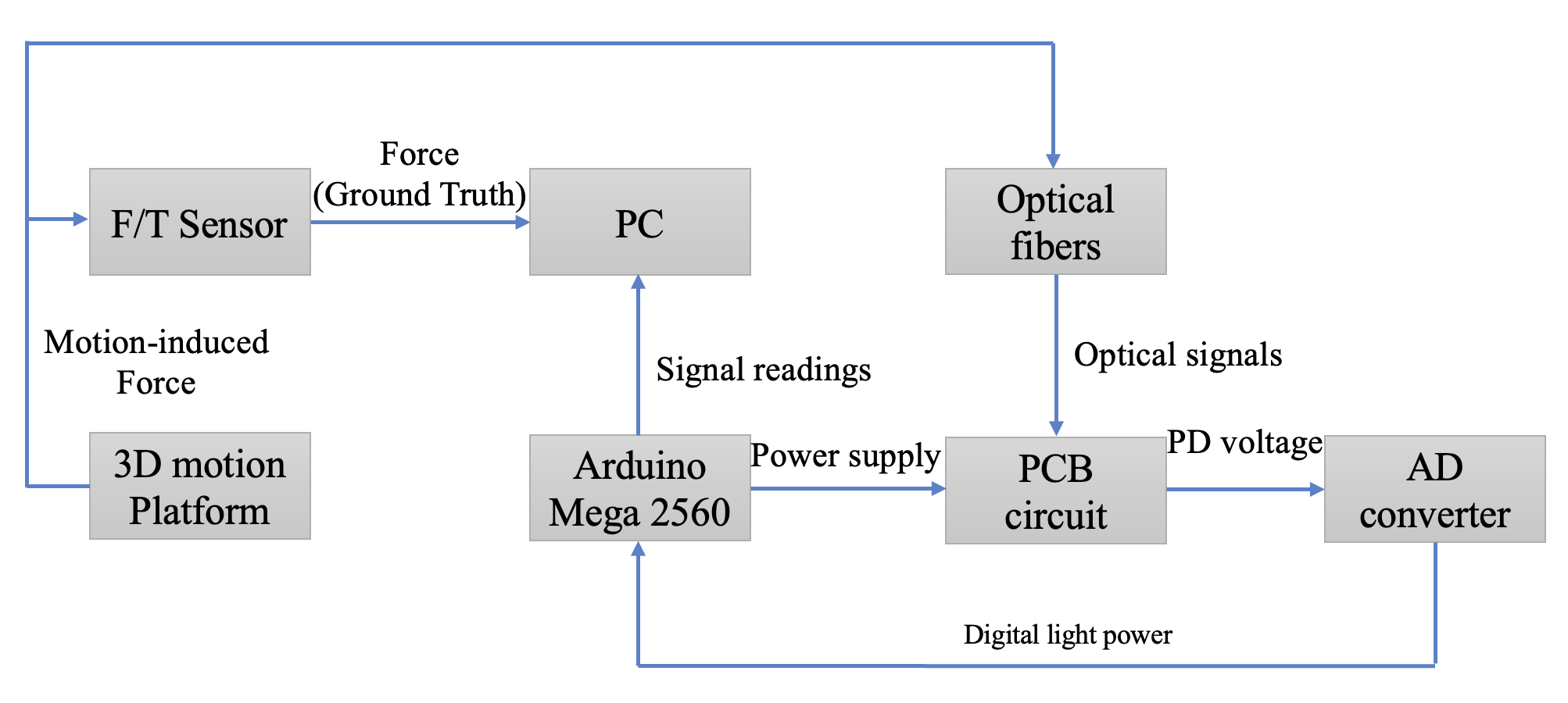}}
    \hfill
    \subfloat[The hardware setup for data acquisition and calibration. The indenter was connected to the F/T sensor and pressed on the surface of the tactile sensor on the motion platform.]{\includegraphics[width = 0.9\columnwidth]{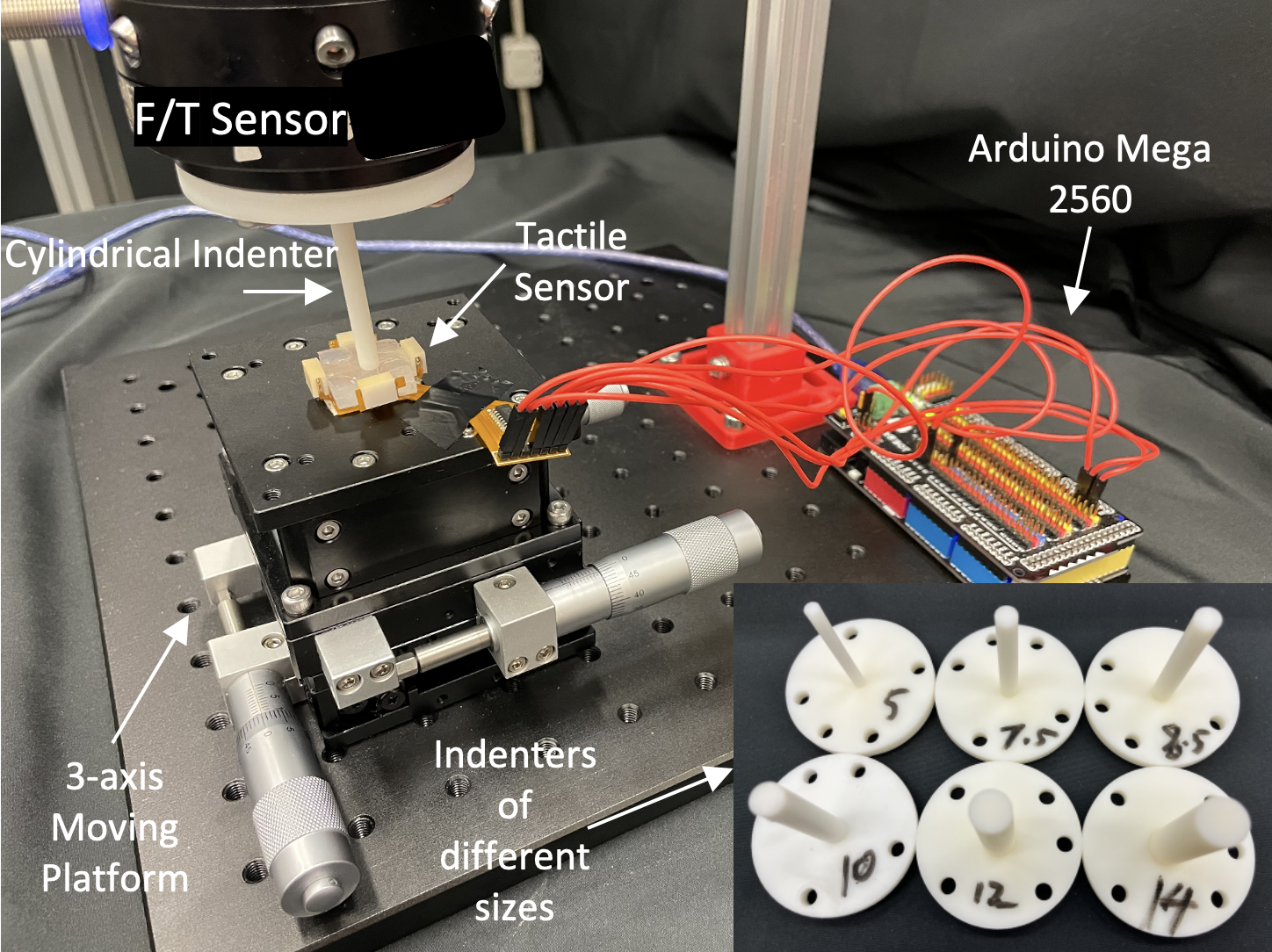}}
    \caption{The diagram and the optical photo of hardware setup for data acquisition and calibration. }
    \label{fig:4}
\end{figure*}

\prettyref{fig:4} depicts the data acquisition process. A 3-axial optical precision motion platform (SZH Co., Ltd.) was used to control the normal and lateral displacement of the sensor and hence how the elastomer was pressed. The maximum moving distance of the motion platform is 8mm in the Z-direction, and 10mm in $\pm$ X- and Y-directions within less than 5$\mu$m repeated location error. In each motion direction, the motion was controlled by manual adjustment. By adjusting the location of the motion platform, relative motion between the platform and the indenter was generated after the contact, along with the induced normal and shear force. The ground truth normal and shear forces are read from the Force/Torque sensor mounted at the tip of a robot arm. 

\subsection{Indentation Depth and Size Sensing} 
As mentioned in Sec.~\ref{sec:principle}, the information of pure indentation in the normal direction is extracted from the lower layer fibers and the bottom PD. Therefore, the responses of the lower layer optical cores, the upper layer optical cores, and the PD at the bottom of the sensor are collected at the displacements of the platform in normal direction ranging from 0-5mm, with cylindrical-shaped indenters with diameters ranging from 5mm to 12mm. The normal force values induced at this point are not for the normal force calibration, but only for generating different indentation depths using indenters of different diameters (as indenters of different sizes generate different indentation depths). Using the previously mentioned hardware setup, the normalized optical intensity changes of ($\Delta I$) divided by $I_0$ reflect the optical fiber deformation and the indentation depth on the sensor. Based on the acquired data, we intend to find the relationship between the readings of the lower layer straight optical fibers and the bottom PD reading and the deformation created purely by the variation of indentation depth and indenter size. The details of calibration can be found in Sec.~\ref{calibration}.

\subsection{Normal and Shear Force Sensing}
As mentioned in Sec.~\ref{sec:principle}, the information of shear force is extracted from the upper layer fibers. To apply shear force onto the sensor, a normal force should also be loaded simultaneously. The indenter was kept vertical to the surface of elastomer during the indentation, and the shear force is generated using the motion of the moving platform in the X- and Y-directions. The different combinations of X- and Y-direction movements generated different shear force components for calibration (see Section \ref{calibration}). To collect individual fiber's response to different values of shear forces, the normal force is kept constant by maintaining the height of the motion platform, and the same process repeats at different normal forces. As the shear force data is collected with the presence of normal forces, the signals of all six optical fibers should be put together for accurate calibration of the relationship between the responses of the optical fibers and the normal and shear forces.

\subsection{Self-Calibration Algorithm} \label{calibration}
Here, inspired by \cite{cao2022polymer}, we proposed a calibration method that determines the normal and shear forces and the indenter's radius from the sensor's optical intensity response. Considering the responses of optical fibers, we use a linear calibration method that calibrates and decouples the normal force, shear force, and the indenter diameter simultaneously. Here, we use a strategy of two-step calibration: (1) calibrate the relationship of light intensity change primarily due to the indentation depth $\delta$ and the radius $r$ and (2) calibrate the relationship of light intensity change primarily due to the applied force. The complete process of self-calibrated measurement of tactile force is presented in Fig.\ref{fig:8}. 

In the first calibration step, we calibrated the linear relationship between the light intensity change sensed by the lower-layer PDs (PD5 and PD6) and the bottom PD7 (three outputs in total) and the indentation factors. We characterize this relationship with the gain matrix $ [R]_{3 \times 2}$ following Equation \ref{eq1}. Each row in $[I_{lower}]_{3 \times n}$ represents one  signal among the $n$ sample signals collected, which contains three PD readings at the lower layer and the bottom. Each row in $[U]_{2\times n}$ contains the concatenated indentation depth and indenter radius of each of the $n$ indentations. 

\begin{equation} \label{eq1}
    [I_{lower}]_{3 \times n} = [R]_{3 \times 2} \cdot [U]_{2\times n}       
\end{equation}

Using the Moore-Penrose pseudoinverse, the solved gain matrix $[\overline{R}]$ is

\begin{equation} \label{eq2}
    [\overline{R}]_{3\times2} = [I_{lower}]_{3\times n} \cdot ([U]^\top_{n\times2} \cdot [U]_{2\times n})^{-1}\cdot [U]^\top_{n \times 2}
\end{equation}

And using this gain matrix, we can recover the indentation depth and indenter diameter simultaneously from the two lower layer PDs and the bottom PD readings by

\begin{equation} \label{eq3}
    [\overline{U}]_{2\times 1} = ([\overline{R}]^\top_{2\times 3} \cdot [\overline{R}]_{3\times2})^{-1} \cdot [\overline{R}]^\top_{2\times3}\cdot [I_{lower}]_{3\times1}
\end{equation}

In the second calibration step, we assume that the total intensity change within the optical fibers is the summation of the intensity change due to the indentation factors calculated in the first step above, and the intensity change due to the force exerted, as shown by

\begin{equation} \label{eq4}
    [I]_{6\times n} = \underbrace{[C]_{6\times 3}\cdot [F]_{3\times n}}_{[I_{force}]_{6\times n}} + \underbrace{[K]_{6\times 2}\cdot[U]_{2\times n}}_{[I_{indent}]_{6\times n}}
\end{equation}
We denote the part of (\ref{eq4}) on the right of the plus sign as $[I_{indent}]$, which contains the $n$ sampled set of signal changes within all six optical fibers when only indentation in the normal direction occurs. The gain matrix $[K]$ calibrates the relationship between the indentation factors and the signal changes in all optical fibers. 

\begin{equation} \label{eq5}
    [I_{indent}]_{6\times n} = [K]_{6\times 2}\cdot[U]_{2\times n}
\end{equation}

The samples in $[I_{indent}]$ are collected with indenters of various sizes at different indentation depths, without the exertion of shear forces. Although normal forces exist during this process, it is both positively correlated with indentation depth and indenter radius, so we will leave the normal force to be considered in the latter equations. Using the Moore-Penrose pseudoinverse as before, the solved gain matrix $[\overline{K}]$ is 

\begin{equation}\label{eq6}
    [\overline{K}]_{6\times2} = [I_{indent}]_{6\times n} \cdot ([U]^\top_{n\times2}\cdot[U]_{2\times n})^{-1} \cdot [U]^\top_{n\times2}
\end{equation}

After $[K]$ is calibrated, the remaining gain matrix to be calibrated is $[C]$, which describes the relationship between the exerted normal and shear forces $[F]$ and the optical signal changes with the indentation factors ignored $[I_{force}]$. The detailed expressions are shown in Eqn. \ref{eq7} and \ref{eq8}.

\begin{equation} \label{eq7}
    [I_{force}]_{6\times n} = [I]_{6\times n} - [I_{indent}]_{6\times n} = [C]_{6\times 3}\cdot [F]_{3\times n}
\end{equation}

\begin{equation} \label{eq8}
    [\overline{C}]_{6\times3} = [I_{force}]_{6\times n}\cdot([F]^\top_{n\times3}\cdot[F]_{3\times n})^{-1} \cdot [F]^\top_{n\times3}
\end{equation}

With all the calibration matrices calculated, the exerted normal and shear forces can be calculated by the calibrated $[C]$ as in Eqn.\ref{eq9}, based on the solved indentation factors $[\overline{U}]_{2\times1}$ and $[\overline{I_{indent}}]_{6\times1}$ fitted from the calibrated $[\overline{K}]_{6\times2}$ and $[\overline{U}]_{2\times1}$. 

\begin{equation} \label{eq9}
    [\overline{F}]_{3\times 1} = ([\overline{C}]^\top_{3\times6}\cdot [\overline{C}]_{6\times3})^{-1} \cdot [\overline{C}]^\top_{3\times 6} \cdot [\overline{I_{force}}]_{6\times 1}
\end{equation}

%% file: sections/experiments.tex
\begin{figure}[t]
    \centering
    \includegraphics[width=\columnwidth]{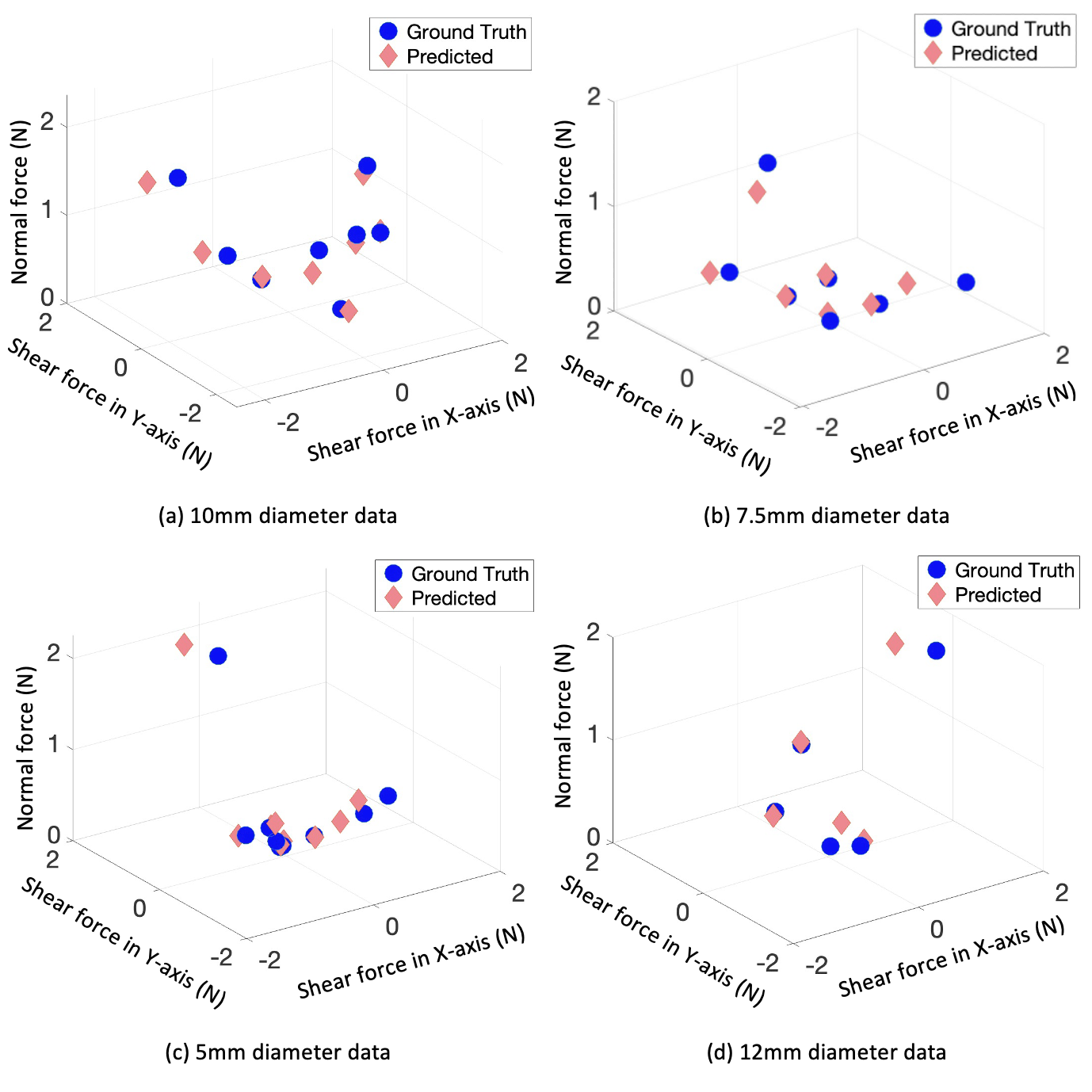}
    \caption{The real values (blue circle) and predicted values (pink diamond) of the normal and shear force combinations exerted by indenters of different diameters. The indenter diameters used in experiments are 10mm, 7.5mm, 5mm, and 12mm, with the corresponding results shown in (a), (b), (c), and (d), respectively. The predicted values are calculated from the signals of 6 optical fibers (PD1-6).}
    \label{fig:5}
\end{figure}

\begin{figure}[t]
    \centering
    \includegraphics[width=1.05\columnwidth]{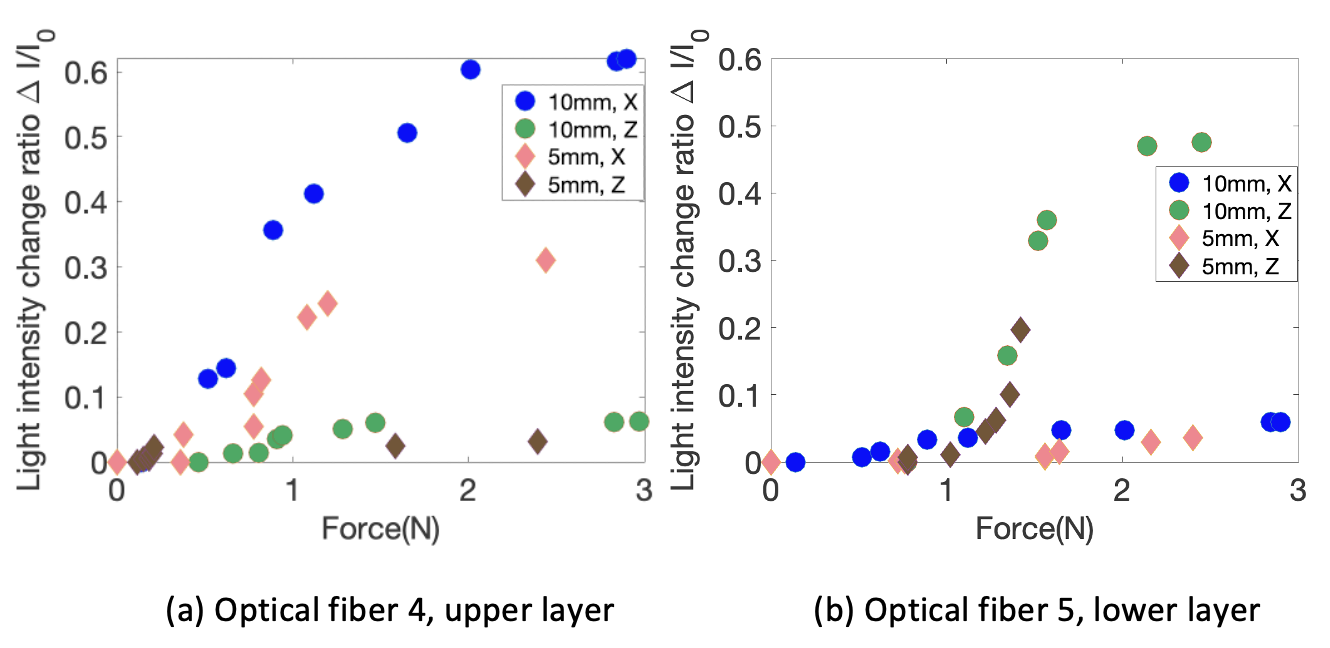}
    \caption{The comparison of responses optical fibers to different indenter diameter sizes and forces in different directions. (a) shows the response of the optical fiber connecting LED4 and PD4, and (b) shows the response of the optical fiber connecting LED5 and PD5. The anisotropicity of each fiber's response is demonstrated.}
    \label{fig:6}
\end{figure}

\begin{figure}[t]
    \centering
    \includegraphics[width=0.7\columnwidth]{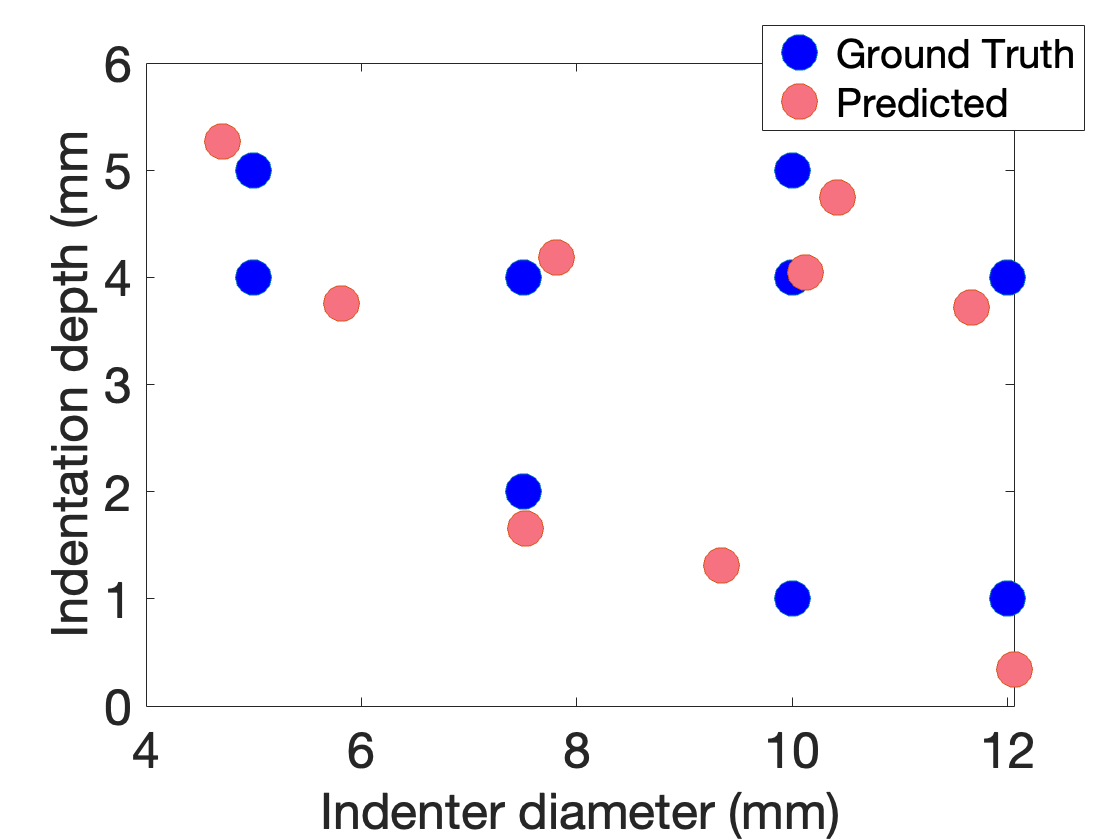}
    \caption{The real values (blue circle) and predicted values (pink circle) of the indentation depth and the indenter radius used in the experiment. The predicted values are calculated from the signals of two optical fibers in the lower layer, and the PD at the bottom (PD5-7).}
    \label{fig:7}
\end{figure}

\section{Results and Discussion}

The comparison between real forces and predicted forces is shown in Fig.~\ref{fig:5}. Based on the linear calibration relationship in Sec.~\ref{calibration}, the predicted force values are calculated from the light signal change, when the motion stage moves to different locations. The motion range used in the experiment in Z-direction is between 0 and 5mm from the initial position, and the motion in $\pm$ X- and Y-direction ranges from 0 and 4mm from the initial position, with the incremental step of 1mm for collecting the calibration dataset. The test dataset was collected using the middle positions between each two positions of the calibration dataset in each direction. The test result is as shown in Fig.~\ref{fig:5}. for each indenter diameter, the predicted force values are closely matched to the real force values within the range of sensing, with the average force error being 0.17N on X-axis, 0.18N on Y-axis, and 0.15N on Z-axis. Although the calibration position of the motion platform was linearly adjusted in each direction, the distribution of the data points corresponding to each calibration position presents nonlinearity, since the elastomer and the embedded optical fiber within present nonlinear strain during the linear compression of the indenter. As the force magnitude increases, this nonlinearity becomes more significant and the data points are distributed more sparsely on the graph. Nevertheless, the linear calibration method can still be used as the force sensing range is still relatively small (within 2N) and the force error is also small compared to the force sensing range. 

As the accuracy of force measurement by our tactile sensor is demonstrated in Fig. \ref{fig:5}, the anisotropic response of optical fibers in different layers is quantitatively shown in Fig. \ref{fig:6}. The response of one of the optical fibers in the upper layer (the fiber connecting LED4 and PD4, see Fig.~\ref{fig2}), is plotted in Fig.~\ref{fig:6}a. The curved optical fiber presents a significant change of light intensity under shear force by indenters of different sizes, while the response of light intensity under normal force can be negligible. The response of one of the optical fibers in the lower layer (the fiber connecting LED5 and PD5, see Fig.~\ref{fig2}), is plotted in Fig.~\ref{fig:6}b. Different from the curved fiber, the straight optical fiber shows a significant change in light intensity under normal compression and negligible change under shear force. The straight fiber also presents significant differences in responses under the normal force by indenters of different diameters. On the other hand, using the anisotropic response of the straight fibers in the lower layer and the response of PD at the bottom, the tactile sensor is able to decouple the indentation depth and the indenter diameter accurately as shown in Fig.~\ref{fig:7}, with an average error of 0.4mm in indenter diameter prediction, and an average error of 0.3mm in indentation depth measurement. The results verify that our design of sensor enables the extraction and decoupling of forces and indenter size simultaneously thanks to the anisotropic responses of optical fibers in different layers. 

%\textcolor{red}{Talk about why choose the current dimension and the shape of the fiber}

%\subsection{Sensitivity and repeatability}
%\begin{itemize}

%\item gauge factor of each channel.
%\item Repeatability, hysteresis, loading and unloading curve

%\end{itemize}

%% file: sections/conclusion.tex
\section{Conclusions}
\label{sec:concl}
In this paper, we have demonstrated a novel polymer-based optical fiber tactile sensor that can measure the contact object size and the decouple normal and shear forces calibrated by the object size, with good accuracy. The results highlight the accurate measurement enabled by the hardware design of the anisotropic two-layered optical fiber structure and flexible PCB, and the calibration algorithm. This relieves the user of tactile sensor from relying on data-driven methods such as machine learning or iFEM in order to accurately measure the deformation field with a contact force and object size information, with relatively low cost and 
ease of fabrication. The texture of the sensor is also conformable and is potentially suitable to be wearable devices for human or robotic skin. 

There are some aspects that can be improved based on our work in the future. Currently, the arrangement of optical fibers, the number of optical fiber layers, and the dimension of the sensor are still limited by the size of the peripheral electronic components. In the future, more layers can be added to decode more tactile modalities, and the different anisotropicity can be adjusted by adopting an optimized arrangement of optical fibers. 
%\textcolor{red}{Talk about what happens if we use multiple layers, and what tests are needed for verification.}